\documentclass[times, review, 10pt]{elsarticle}

\usepackage{amssymb}

\usepackage{amsmath}

\usepackage{algorithm}
\usepackage{algpseudocode}
\usepackage{algorithmicx}
\usepackage{float}
\usepackage{array}

\usepackage{textcomp}
\usepackage{stfloats}
\usepackage{verbatim}
\usepackage{booktabs}
\usepackage{wrapfig}

\usepackage{animate}
\usepackage{media9}

\usepackage{xcolor}
\usepackage{colortbl}
\usepackage{makecell}
\usepackage{multirow}
\usepackage{wrapfig}
\usepackage{caption}
\usepackage{hyperref}

\newcommand{\eg}{\textit{e}.\textit{g}.}

\newcommand{\etal}{\textit{et~al.}}

\begin{document}

\begin{frontmatter}

\title{Every Painting Awakened: A Training-free Framework for Painting-to-Animation Generation}

\author[lly]{Lingyu Liu} 
\affiliation[lly]{organization={School of Software Engineering, Xi'an Jiaotong University},
            country={China}}
\ead{liulingyu@stu.xjtu.edu.cn}
\author[wyx]{Yaxiong Wang\corref{cor1}} 
\affiliation[wyx]{organization={School of Computer and Information Science, Hefei University of Technology},
            country={China}}
\ead{wangyx15@stu.xjtu.edu.cn}
\author[lly]{Li Zhu} 
\ead{zhuli@xjtu.edu.cn}
\author[zzd]{Zhedong Zheng\corref{cor1}} 
\affiliation[zzd]{organization={Faculty of Science and Technology, University of Macau},
            country={China}}
\ead{zhedongzheng@um.edu.mo}
\cortext[cor1]{Corresponding author.}

\begin{abstract}
We introduce a training-free framework specifically designed to bring real-world static paintings to life through image-to-video (I2V) synthesis, addressing the persistent challenge of aligning these motions with textual guidance while preserving fidelity to the original artworks. 
Existing I2V methods, primarily trained on natural video datasets, often struggle to generate dynamic outputs from static paintings. 
It remains challenging to generate motion while maintaining visual consistency with real-world paintings. This results in two distinct failure modes: either static outputs due to limited text-based motion interpretation or distorted dynamics caused by inadequate alignment with real-world artistic styles.
We leverage the advanced text-image alignment capabilities of pre-trained image models to guide the animation process. Our approach introduces synthetic proxy images through two key innovations: 
(1) \textbf{Dual-path score distillation:} We employ a dual-path architecture to distill motion priors from both real and synthetic data, preserving static details from the original painting while learning dynamic characteristics from synthetic frames.
(2) \textbf{Hybrid latent fusion:} We integrate hybrid features extracted from real paintings and synthetic proxy images via spherical linear interpolation in the latent space, ensuring smooth transitions and enhancing temporal consistency.
Experimental evaluations confirm that our approach significantly improves semantic alignment with text prompts while faithfully preserving the unique characteristics and integrity of the original paintings. Crucially, by achieving enhanced dynamic effects without requiring any model training or learnable parameters, our framework enables plug-and-play integration with existing I2V methods, making it an ideal solution for animating real-world paintings. More animated examples can be found on our project website \footnote{\url{https://painting-animation.github.io/animation/}}.
\end{abstract}

\begin{keyword}
Artistic Animation Generation \sep Training-free Framework \sep Latent Feature Interpolation \sep Text-Guided Motion Synthesis
\end{keyword}

\end{frontmatter}

\section{Introduction}
With advances in diffusion models for generating high-fidelity images, there is a growing focus within the research community on extending these models to video generation tasks. Recently, some works~\cite{seine,mpd,li2024generative} have made significant progress in the field of image-to-video (I2V) generation, enabling the synthesis of dynamic video content from provided static images and text descriptions. These technologies can generate videos with temporal and spatial coherence~\cite{suo2024jointly}, greatly enriching the methods of multimedia content creation and broadening the horizons of visual expression. 
Nonetheless, when utilizing existing I2V models for animating static paintings in the real world, we encounter significant limitations. Specifically, these models frequently struggle to accurately interpret the instructional guidance provided by text prompts, leading to a failure in generating the expected motion effects. Mainstream I2V models are primarily based on natural scene video datasets (\eg, WebVid-10M~\cite{webvid}), lacking the capability to encode artistic features of real paintings and struggling to extract actionable motion latent representations.

\begin{figure}[!t]
    \centering
    \includegraphics[width=0.98\linewidth]{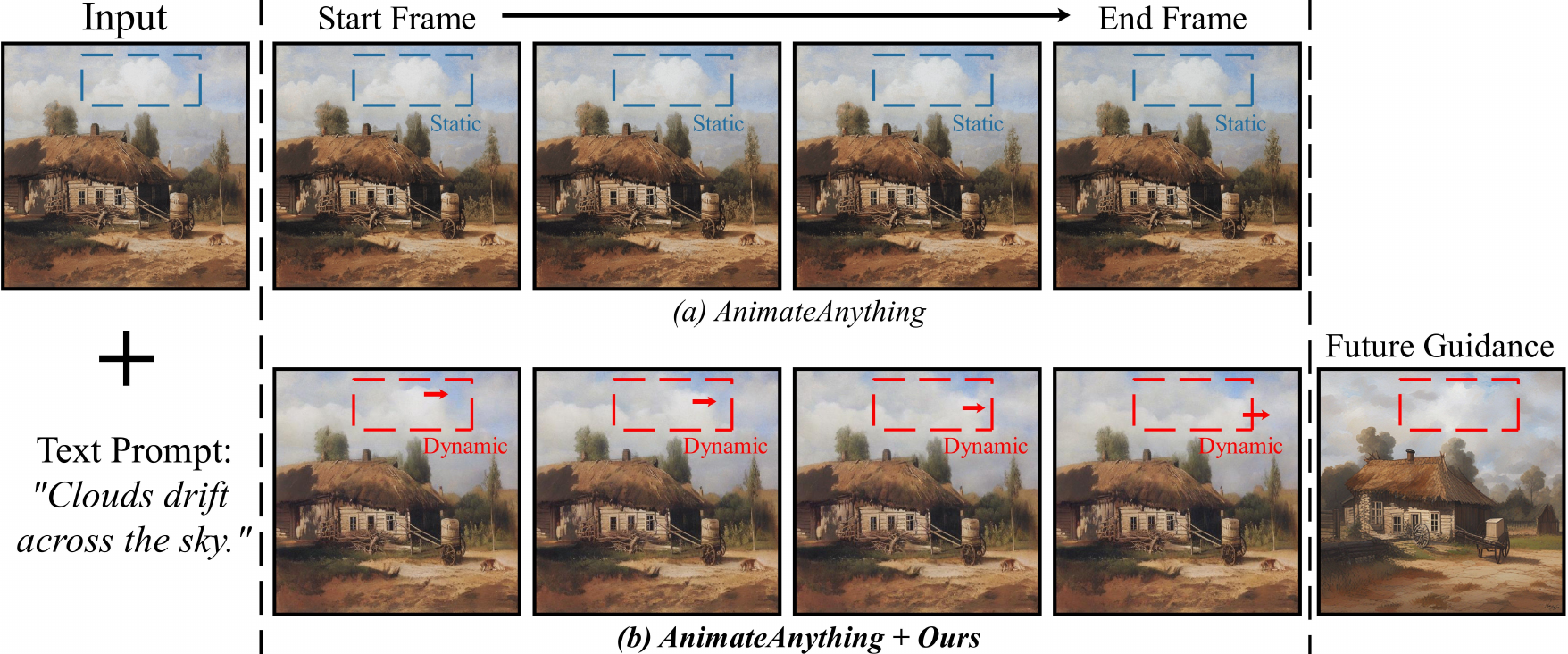}
    \caption{The animation of AnimateAnything (top) vs. our method (bottom).
    Given the prompt \textit{``Clouds drift across the sky"} and a consistent global seed, our method, which utilizes a refined synthetic proxy image for future guidance, shows a significantly improved response to the prompt compared to the base I2V model.}
    \label{fig:ani}
\end{figure}

As illustrated in the top row of Fig.~\ref{fig:ani}, we harness a real painting from the WikiArt dataset~\cite{WikiArt} into a base I2V model (AnimateAnything~\cite{animateanything}), with the text prompt specified as \textit{``Clouds drift across the sky"}. The model encounters challenges in accurately interpreting the meaning of prompts due to its difficulty in understanding the artistic style of the images. Although the output exhibits dynamic effects, it fails to fully realize the motion intent conveyed by the prompt. For instance, the model can only animate the trees in the painting based on the prior knowledge it learned from the training dataset and cannot make the clouds in the sky move according to the instructions in the text prompt. This limitation highlights the need for a more robust framework that can both interpret text prompts accurately and preserve the artistic style of the input.

To address this challenge, we introduce an auxiliary future image, aiming to highlight the textual intent in the image domain and enhance the coherence simultaneously.  
Specifically, we propose synthesizing proxy images for future guidance via pre-trained image diffusion models, which are then integrated into the video diffusion process. 
These image models are trained on extensive image-text paired datasets, enabling them to respond effectively to text prompts and enhance low-quality images to high-quality versions.
As illustrated in the bottom row of Fig.~\ref{fig:ani}, we obtain a new high-quality synthetic proxy image by feeding the painting along with the text prompt into an image refinement model (SDXL Refiner~\cite{sdxl}). This proxy image maintains a similar structure to the original, with the subjects within it more clearly defined according to the text prompt. 
With the same global seed maintained for generating real painting animations, the incorporation of synthetic proxy images enables the video model to accurately identify the sky region and correctly interpret the motion intent contained in the text prompt.

\begin{figure}[!thbp]
    \centering
    \includegraphics[width=1\linewidth]{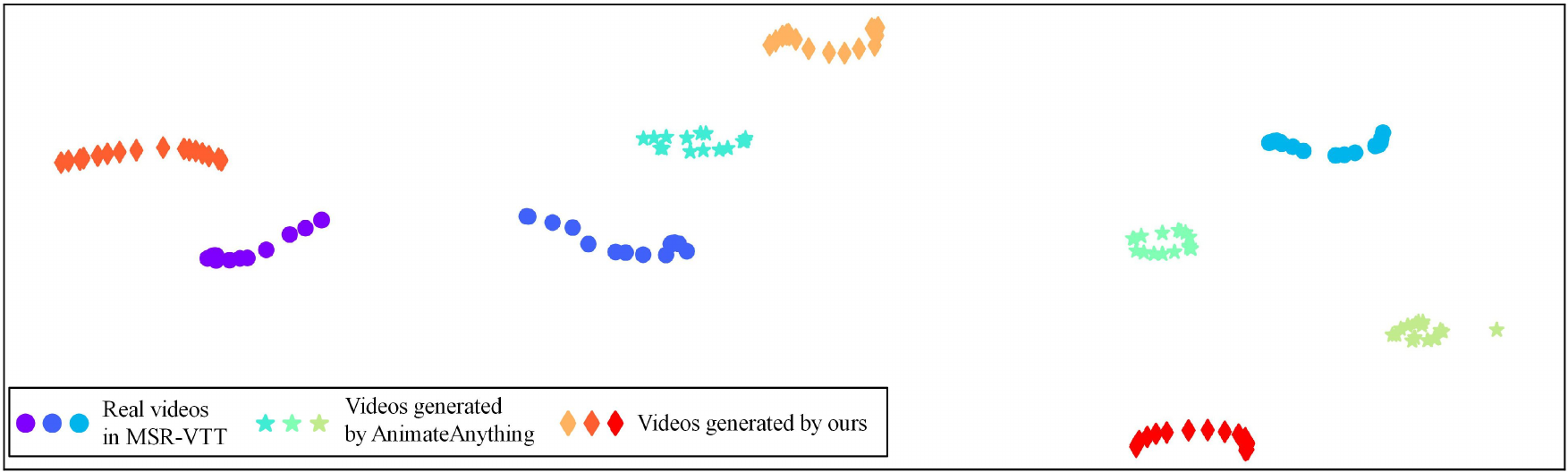}
    \caption{Visualization of different video latent vectors. We employ t-SNE to visualize the latent vectors of different videos in the feature space,
    where each point represents a video frame. Frames from the same video form a single cluster. Clusters formed by real videos from the MSR-VTT dataset exhibit a linear trend, indicating that frames of real videos are orderly arranged in their feature space. In contrast, clusters formed by videos synthesized using AnimateAnything are tightly packed and do not show a clear single linear trend. Clusters generated by videos recreated using our method share a similar linear trend with those of the real videos. This suggests that our approach achieves high fidelity and continuity in capturing video features.}
    \label{fig:tSNE}
\end{figure}

Furthermore, to more smoothly integrate the information from synthetic proxy images into the video diffusion model, we introduce the score distillation sampling strategy and latent feature interpolation into the image-to-video synthesis.
In our method, we first synthesize the proxy image based on the real painting and text prompt.
Then, we perform dual-path video score distillation sampling on the latent features corresponding to both the real painting and synthetic proxy image, extracting motion priors from the video diffusion model. Following this, we apply spherical linear interpolation (Slerp) along the temporal dimension to the refined latent features of both images, resulting in a new set of latent features. This newly formed latent feature is then fed into the base I2V model to further refine the generation process. Consequently, the produced videos not only retain the informational essence of the original image but also align with the expected motion effects dictated by the text prompt. Through this approach, we ensure an enhanced output video that well expresses the textual instruction and excels in terms of motion consistency. 

We present the t-SNE visualizations of the feature representations of frames extracted from both real and synthetic videos in Fig.~\ref{fig:tSNE}, where each point represents a video frame. The real video samples are sourced from the MSR-VTT dataset~\cite{msr}. We observe that frames from the same video form a single cluster. The clusters formed by real videos exhibit a clear linear trend, indicating their coherence in both time and space. In contrast, clusters formed by synthetic videos generated by AnimateAnything~\cite{animateanything} are densely packed and lack a distinct linear pattern. However, clusters formed by synthetic videos recreated using our method show a similar distribution to those of real videos, suggesting that our approach effectively captures the temporal and spatial characteristics of real videos. Specifically, our contributions in this work can be summarized as follows:
\begin{itemize}
\item We introduce a new inference framework that harnesses pre-trained image-to-video models to animate real paintings effectively. By integrating synthetic proxy images, which are optimized via an image diffusion model, into the inference pipeline, our method significantly improves the generation quality of the resulting video sequences. Notably, this enhancement is achieved without requiring additional training or introducing new learnable parameters, allowing for seamless integration with current I2V models in a plug-and-play manner.

\item We employ a dual-path video score distillation sampling strategy to independently refine the initial latent features corresponding to real paintings and synthetic proxy images. Subsequently, spherical linear interpolation is applied between these refined latent features to derive a hybrid latent feature. The frame cluster distribution of videos generated from the hybrid latent feature aligns more closely with that of real-world videos, enhancing the fidelity and realism of synthesized video content.

\item We assess the performance of our method through detailed experiments. Evaluation results indicate that our method effectively generates videos that not only retain the informational fidelity of the original images but also exhibit motion dynamics that accurately reflect the intent specified by the text prompts.
\end{itemize}

\section{Related Work}
\subsection{Diffusion Models for Image Generation}
Compared to traditional methods based on GANs~\cite{gan,pal2025synthesizing,zhang2023large,zhou2024adaptive} and VAEs~\cite{vae,sun2024reparameterizing}, diffusion models~\cite{denoising,song2025attridiffuser,liu2025textdiff}, as an emerging generative model technique, have 
achieved high-quality image generation by progressively adding noise to the data and then learning the reverse process to recover it from the noise. Recent studies have shown that diffusion models possess unique advantages in handling complex image generation tasks~\cite{sd_high,glide,Imagen,controlnet}. The SDXL Refiner~\cite{sdxl}, the core optimization component of Stable Diffusion-XL, can receive both text and image inputs to enhance the quality of the input images through its optimization process. InvSR~\cite{invsr} is an open-source image super-resolution tool based on the inverse process of diffusion models. It aims to efficiently enhance the clarity of low-resolution images using the image prior knowledge from pre-trained models. Qiao~\etal~\cite{qiao2025learning} introduce PhyDiff, a model enhancing zero-shot image restoration by integrating physical priors with diffusion models for superior image generation, while Zeng~\etal~\cite{zeng2024instilling} involve the multi-round consistency for the long-term image editing. InstructPix2Pix~\cite{instructpix2pix} is a natural language instruction-based image editing model that integrates natural language understanding with image generation. MasaCtrl~\cite{masactrl} focuses on achieving consistency in image synthesis and editing without the need for fine-tuning. Building on the advancements of these image models, our work introduces synthetic proxy images to enhance the effectiveness of generating image animations.

\subsection{Diffusion Models for Video Generation}
In the case of current I2V models~\cite{gid,mim,conditional} in the open domain, these primarily leverage a pretrained text-to-video model as a foundation, relying on supervision that enhances image motion information to guide the video diffusion models. Compared to the text-to-video task~\cite{animatediff}, the image-to-video task adds an additional requirement to maintain the original style of the input image.
VideoComposer~\cite{videocomposer} and Dynamicrafter~\cite{dynamicrafter} leverage the CLIP embedding of the image as an additional input to retrain the text-to-video diffusion model, aiming to generate dynamic videos related to the input image. Nevertheless, the features extracted by the CLIP image encoder generally ignore the local information of the original image, which results in the generated videos not being entirely consistent with the original image. 
ConsistI2V~\cite{consisti2v} and Cinemo~\cite{cinemo} deploy a lightweight network to preserve the detailed features of the input image. ConsistI2V~\cite{consisti2v} applies a cross-frame attention mechanism to input the features of the input image into the spatial layers of the diffusion model, thereby retaining the information of the original image while generating the video. Cinemo~\cite{cinemo} trains the text-to-video model to predict the motion residual distribution corresponding to the input image, thereby achieving dynamic effects for the image. Despite improving the visual alignment with the original image, these methods do not adequately understand the input text prompt and tend to generate static videos.
To effectively regulate the intensity of moving objects in image animations, AnimateAnything~\cite{animateanything}, the animation branch of VideoComposer~\cite{videocomposer}, incorporates a motion intensity metric during the training phase. While this method yields superior dynamic effects in images, the videos generated by this approach predominantly rely on the motion priors embedded within the training set, resulting in a less responsive adaptation to provided text prompts. Our method enhances the generative capabilities of these approaches by incorporating information from synthetic proxy images, specifically tailored for animating real paintings.

\section{Every Painting Awaken Framework}
\begin{figure}[!t]
    \centering
    \includegraphics[width=0.98\linewidth]{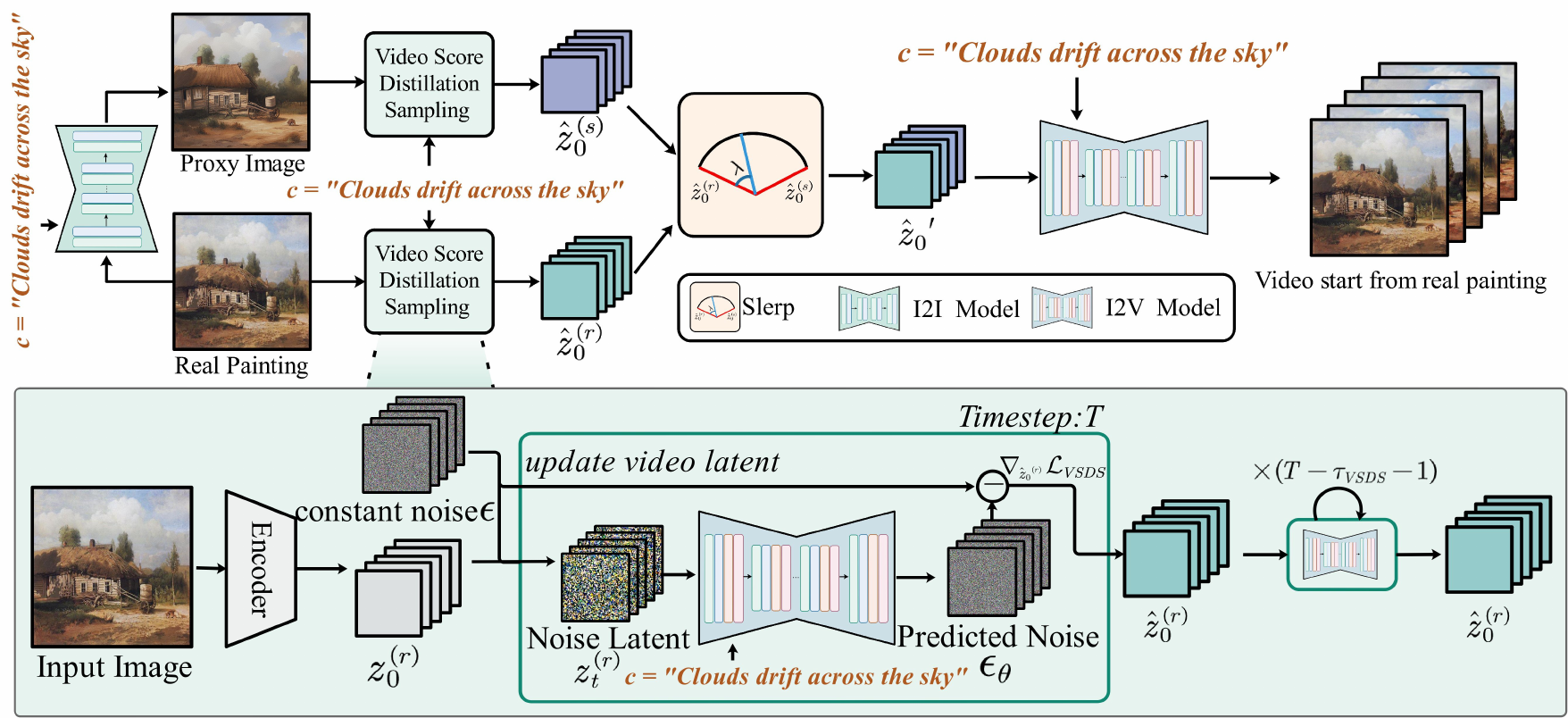}
    \caption{Illustration of our method. Given a real painting and its synthetic counterpart refined by an image diffusion model, we apply dual-path video score distillation sampling to infuse their latent vectors with motion information. Next, we perform spherical linear interpolation on these updated latent vectors across the frame dimension. The hybrid latent vectors are then fed into the I2V model to generate dynamic videos.
    }
    \label{fig:framework}
\end{figure}
\noindent \textbf{Overview.} The entire process of our method is illustrated in Fig.~\ref{fig:framework}.
We first apply a pre-trained image refinement model to generate a synthetic proxy image $I_s$
from the real painting $I_r$, with $I_s$ serving as future guidance for subsequent steps.
We apply dual-path video score distillation sampling to the real painting and the synthetic proxy image, obtaining two updated initial video latent vectors ${\hat{z}_0}^{\left( r \right)}$ and ${\hat{z}_0}^{\left( s \right)}$. These vectors are then spherically interpolated along the temporal dimension to generate a fused latent vector ${\hat{z}_0}^{\prime}$. This fused vector is subsequently used as input to the I2V model for video generation.
To better understand our approach, we first introduce the basic paradigm of video generation models in Sec.~\ref{sec:pre}. 
Subsequently, we present detailed descriptions of each component: Sec.~\ref{sec:syn} covers the synthesis of proxy images; Sec.~\ref{sec:dis} discusses dual-path score distillation; and Sec.~\ref{sec:inter} explains hybrid feature fusion.

\subsection{Preliminary: Video Latent Diffusion Model}
\label{sec:pre}
The video latent diffusion model (VLDM) is an extension of the image latent diffusion model (ILDM) along the temporal dimension, designed to handle three-dimensional video data. We mainly introduce the basic structure and fundamental knowledge of the image-to-video diffusion model.
Formally, in training process, VLDM first leverages a VAE encoder to convert each video frame into a lower-dimensional latent vector $z_0\in \mathbb{R} ^{L\times C\times H\times W}$, where $L$ is the number of frames in the video, $C$, $H$, and $W$ represent the number of channels, height, and width of the latent vector, respectively. Within this lower-dimensional latent space, the diffusion process is formalized as $z_t=q\left( z_t|z_{t-1} \right) $, where $t\in \left[ 1,T \right] $ is the discrete time step index, and $q$ is the probability distribution of the noise addition process. VLDM trains a denoising model $p$ to predict the preceding latent state $z_{t-1}$ from a given noisy latent state $z_t$. The training of this model involves minimizing a loss function defined as follows:
\begin{equation}
    \underset{\theta}{\min}\mathbb{E} _{z_0,t,\epsilon}\left[ \left\| \epsilon -\epsilon _{\theta}\left( z_t,c,t \right) \right\| _{2}^{2} \right] ,
\end{equation}
where $\epsilon$ and $\epsilon_{\theta}$ represent the actual noise and the predicted noise respectively, and $c$ denotes the text, image, or other conditioning factors. During the inference phase of the image-to-video task, the process begins by replicating the original image to be animated to match the required number of frames for the target video length, thus constructing an initial static video sequence. Subsequently, the VLDM enriches the static sequence by integrating dynamic elements that are embedded within the conditioning factors, thereby effectively introducing motion characteristics into the final video.

\subsection{Future Guidance Synthesis}
\label{sec:syn}
Given a real painting $I_r$ and the text prompt, we utilize a text-driven image diffusion model, such as SDXL refiner~\cite{sdxl}, to produce a refined synthetic representation $I_s$ of the original artwork. Leveraging the advanced capabilities of the well-trained text-to-image model, the synthetic image is not only well-aligned with the text prompts but also of high quality, serving as the visual intention proxy for the text instruction in the image domain. Throughout the generation process, the prompt supplied to the image model is consistent with the prompt applied for subsequent animation. This strategy ensures that the proxy image effectively incorporates the intended motion in the visual domain, thereby guiding the video models to generate animations that meet the expectations of the text prompt while maintaining coherence.

\subsection{Dual-path Score Distillation}
\label{sec:dis}
Current mainstream I2V methods typically replicate the initial image during inference to construct an initial static video sequence, which is subsequently utilized as input for the model. This process introduces a significant discrepancy between the training phase, which is grounded in dynamic real-world video data, and the inference phase, where the model is initialized with static inputs. Furthermore, these models tend to apply motion effects based on motion priors learned from the training dataset, rather than accurately interpreting and responding to the specific instructions provided by the text prompt.
Our method incorporates synthetic proxy images into the animation of real paintings, ensuring that the generated videos achieve the motion effects desired by the text prompt. We first perform video score distillation sampling on both the real painting and the synthetic proxy image separately to obtain their dynamic latent vectors. This approach helps to bridge the gap between training and inference to a certain extent, ensuring that the generated videos are more consistent with real-world dynamics.

Score distillation sampling~\cite{dreamfusion} employs the concept of score functions, which are the gradients of the log-probability density function, to effectively distill knowledge from a teacher model to a student sampler.
I4VGEN~\cite{i4vgen} extends this technique to the text-to-video domain by designing a video score distillation sampling to acquire motion priors from the VLDM. In video score distillation sampling, they directly optimize parameterized representations of static videos. 
During the optimization process, Gaussian noise $\epsilon$ is sampled only once to maintain consistency across iterations, which in turn accelerates the convergence speed. The loss function is defined as:
\begin{equation}
   \nabla _{\hat{z}_0}\mathcal{L} _{VSDS}=\omega \left( t \right) \left( \epsilon _{\theta}\left( z_t,c,t \right) -\epsilon \right),
    \label{eq:vsds}
\end{equation}
where $z_t$ represents the latent vector of the noisy video at time step $t$. $\omega \left( t \right)$ represents the weight assigned to the noise component at time step $t$, reflecting how much noise is present in the input data at the current time step. 

Specifically, as shown in Fig.~\ref{fig:framework}, we employ a VAE encoder to extract the latent representation of the real painting $I_r$ and replicate this latent vector across the frame dimension to obtain the initial latent vector representation of the static video $z_{0}^{\left( r \right)}\in \mathbb{R} ^{L\times C\times H\times W}$, where $L$ represents the number of frames in the generated video. In contrast to traditional score distillation sampling, video score distillation sampling leverages a single sample of Gaussian noise $\epsilon$ from a standard normal distribution, ensuring consistent noise characteristics throughout the process. For each time step $t\in \left[ T,\tau _{VSDS} \right] $, we add this noise $\epsilon$ to the static video vectors to obtain the noise latent $z_{t}^{\left( r \right)}$. The gradient $\nabla _{{\hat{z}_0}^{\left( r \right)}}\mathcal{L} _{VSDS}$ is calculated according to Eq.~\eqref{eq:vsds}. We update ${\hat{z}_0}^{\left( r \right)}$ based on the gradient as follows:
\begin{equation}
   {\hat{z}_0}^{\left( r \right)}\gets {\hat{z}_0}^{\left( r \right)}-\alpha \cdot \nabla _{{\hat{z}_0}^{\left( r \right)}}\mathcal{L} _{VSDS},
   \label{eq:grad}
\end{equation}
where $\alpha$ is a constant weight.
Notably, video score distillation sampling is applied only during the initial stages of the denoising process $\left[ T,\tau _{VSDS} \right]$, where $\tau _{VSDS}=T\times p$ and $p$ denotes a proportionality factor. Following a similar procedure, we obtain the updated latent vector of the synthetic proxy image ${\hat{z}_0}^{\left( s \right)}$. 
We perform distillation sampling on real paintings to maintain content fidelity and on synthetic proxy images to leverage dynamic priors. This process generates two distinct dynamic vectors: ${\hat{z}_0}^{\left( r \right)}$, a content-preserving dynamic vector that ensures temporal continuity by retaining the intrinsic characteristics of the original content across time, providing a coherent sequence faithful to the source material; and ${\hat{z}_0}^{\left( s \right)}$, a text-driven dynamic vector emphasizing semantic conformity, which aligns closely with textual descriptions to ensure that the generated sequences accurately reflect the intended semantics specified by the text prompts.

\subsection{Hybrid Latent Fusion}
\label{sec:inter}
By incorporating the information from the synthetic proxy image into the video generation process, we enable the VLDM to access more image information, thereby making motion predictions that better align with the text prompt instructions. The most straightforward method is to perform uniform linear interpolation of the latent vectors of the real painting and the synthetic image across the frame dimension. However, as Zheng~\etal~\cite{noisediffusion} highlight in their study of image interpolation, in high-dimensional spaces, random variables following a standard normal distribution are primarily distributed on a hypersphere centered around the mean. During the training process, the model can only observe noisy images that predominantly lie on this hypersphere, hence it can effectively restore only those images that exhibit such characteristics.
In video generation, the application of uniform linear interpolation results in increased estimation errors and lower fitting accuracy.
Inspired by Zheng~\etal~\cite{noisediffusion}, we find that performing spherical linear interpolation of the latent vectors across the frame dimension allows us to introduce motion information from the synthetic proxy image while largely preserving the detail information of the original painting.
Formally, we perform uniform spherical linear interpolation of ${\hat{z}_0}^{\left( r \right)}$ and ${\hat{z}_0}^{\left( s \right)}$ across the frame dimension. For each frame $l\in \left[ 0,L-1 \right]$, we calculate a new latent vector $\hat{z_l}^{\prime}$:
\begin{equation}
     \hat{z_l}^{\prime}=\frac{\sin \left( \left( 1-\beta _l \right) \theta \right)}{\sin \left( \theta \right)}\hat{z}_{l}^{\left( r \right)}+\frac{\sin \left( \beta _l\theta \right)}{\sin \left( \theta \right)}\hat{z}_{l}^{\left( s \right)}
\end{equation}
where $\hat{z}_{l}^{\left( r \right)}$ and $\hat{z}_{l}^{\left( s \right)}$ are the latent vectors of the $l$-th frame from ${\hat{z}_0}^{\left( r \right)}$ and ${\hat{z}_0}^{\left( s \right)}$, respectively, $\theta$ is the angle between vectors ${\hat{z}_0}^{\left( r \right)}$ and ${\hat{z}_0}^{\left( s \right)}$, and $\beta _l=\,\,\small{\frac{l}{L-1}}$. We combine all the $\{\hat{z_l}^{\prime}\}_{l=0}^{L-1}$ into a new sequence of latent vectors ${\hat{z}_0}^{\prime}$, which serves as the new input to the I2V model. This method ensures a smooth transition of latent vectors across the frame dimension while maintaining consistency and coherence between frames.

\section{Experiments}

\subsection{Experimental Setup}
\noindent \textbf{Implementation details.} We propose a new image-to-video inference pipeline that is designed with plug-and-play capabilities, enabling its seamless integration into existing image-to-video models. To illustrate the broad applicability and effectiveness of our proposed method, we integrate it into three state-of-the-art image-to-video diffusion models: AnimateAnything~\cite{animateanything}, ConsistI2V~\cite{consisti2v} and Cinemo~\cite{cinemo}. For AnimateAnything and ConsistI2V, we generate videos at a resolution of $512 \times 512$ pixels. For Cinemo, the video generation is performed at a resolution of $512 \times 320$ pixels. For each method, we generate a sequence of 16 consecutive frames to construct a video clip for evaluation purposes, employing the default configurations specified in their respective open-source implementations. To avoid bias due to random factors, we fix the random seed to 42 for each model during testing. We adopt the SDXL refiner~\cite{sdxl} as our image enhancement tool to generate synthetic proxy images. We affirm the effectiveness of our method by showcasing animations generated from real paintings in the WikiArt dataset~\cite{WikiArt}. To more intuitively highlight the efficacy of our proposed method, we construct a test set by randomly selecting 1,000 video-text pairs from the MSR-VTT dataset~\cite{msr}, an established benchmark for open-domain video retrieval.

\noindent \textbf{Evaluation Metrics.} Consistent with previous image-to-video methods~\cite{li2024generative,cinemo,consisti2v}, we utilize the Fréchet Video Distance (FVD)~\cite{fvd} as a quantitative measure to evaluate the quality of the generated video sequences. Furthermore, we apply CLIP similarity (CLIPSIM~\cite{wu2021godiva}) to assess the semantic alignment between the generated videos and their associated text prompts. This dual evaluation allows for a comprehensive analysis of the model's fidelity in generating videos and its responsiveness to text prompts.

\subsection{Comparisons with Competitive Methods}

\begin{table}[!t]
  \centering
  \caption{Quantitative comparison results on the MSR-VTT dataset. ``Baseline" refers to the results of various models tested on the dataset using their default settings as provided in their open-source code. ``+Ours" denotes the test results of each model after incorporating our full inference method. ``+ V" represents the scenario where synthetic proxy images are not introduced, and only video score distillation sampling is performed during inference. ``+ S" indicates the introduction of synthetic proxy images followed by spherical linear interpolation. ``+ V\& U" indicates replacing the spherical linear interpolation in our method with uniform linear interpolation. Bold text is used to highlight the optimal outcomes, whereas underlined text signifies the second-best performance.}
  \resizebox{0.98\linewidth}{!}{
    \begin{tabular}{l|ccc||cc||cc||cc}
    \toprule
    \multicolumn{4}{c||}{Method}  & \multicolumn{2}{c||}{AnimateAnything} & \multicolumn{2}{c||}{ConsistI2V} & \multicolumn{2}{c}{Cinemo} \\
    \midrule
          & VSDS  & Slerp & Uniform & CLIPSIM$\uparrow$ & FVD$\downarrow$   & CLIPSIM$\uparrow$ & FVD$\downarrow$   & CLIPSIM$\uparrow$ & FVD$\downarrow$ \\
\cmidrule{2-10}    Baseline  &       &       &       & 25.79 & 530   & 25.27 & 486   & 24.99 & 614 \\
    \midrule
    +V    & \checkmark &       &       & 25.87 & 420   & 25.66 & 320   & 25.22 & \underline{529} \\
    +S    &       & \checkmark &       & 26.07 & 439   & 25.59 & 309   & 25.00    & 572 \\
    +V\&U & \checkmark &       & \checkmark & \underline{26.28} & \underline{392}   & \underline{25.68} & \underline{308}   & \underline{25.24} & 540 \\
    \midrule
    +V\& S (\textbf{Ours}) & \checkmark & \checkmark &       & \textbf{26.35} & \textbf{317}   & \textbf{25.71} & \textbf{298}   & \textbf{25.33} & \textbf{526} \\
    \bottomrule
    \end{tabular}%
    }
  \label{tab:t1}%
\end{table}%

\begin{figure}[htbp]
    \centering
    \includegraphics[width=0.98\linewidth]{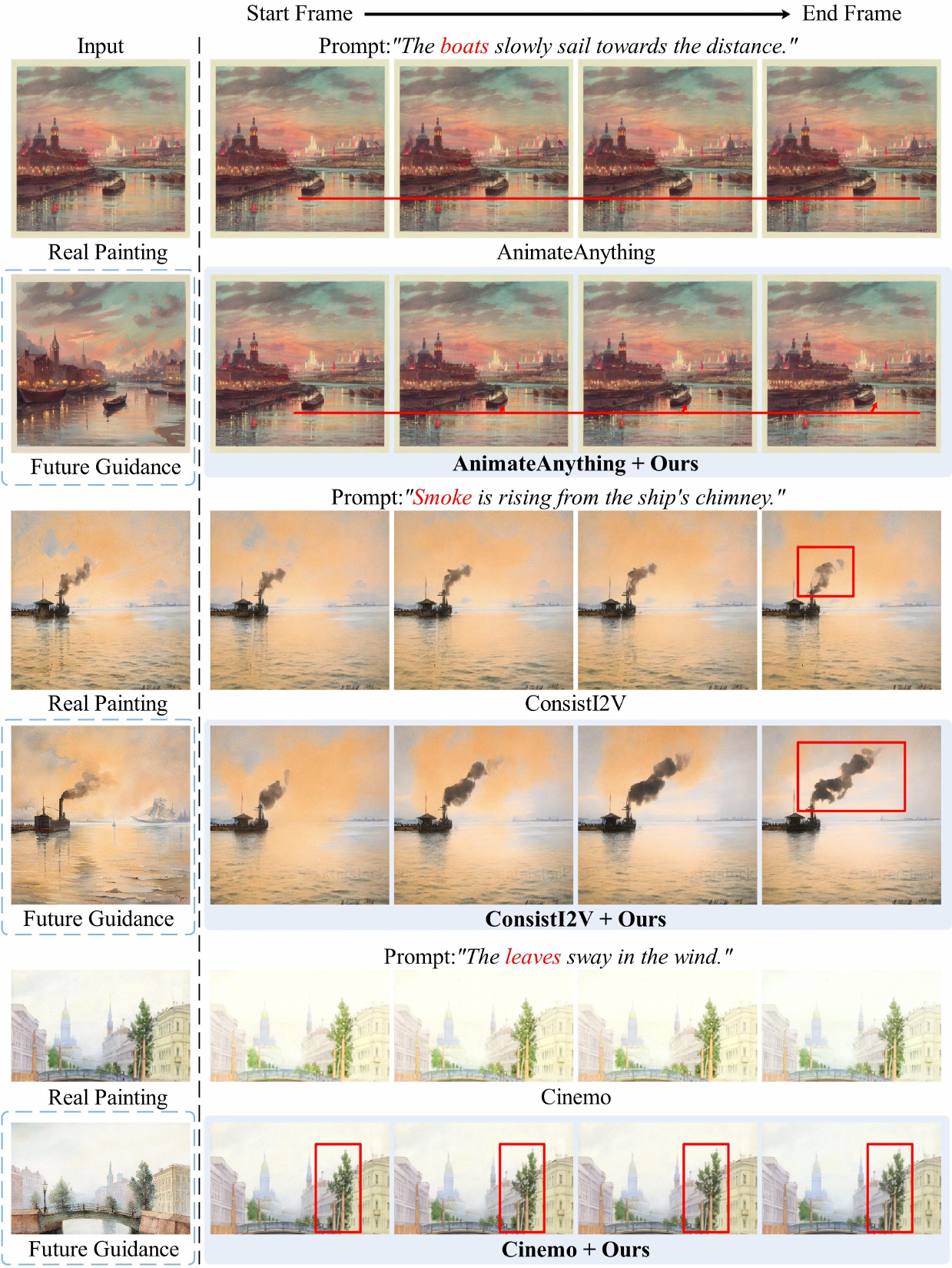}
    \caption{Qualitative comparisons with current image-to-video methods. AnimateAnything fails to interpret the text prompt's request for the ``\textit{boat}” to move, while our method can generate a video showing the boat moving forward. Compared to ConsistI2V, our approach produces more pronounced and superior motion effects for ``\textit{smoke}”. Cinemo struggles with both retaining the input image's information and understanding the prompt's intent. By incorporating our inference method, it not only preserves the input image's style but also successfully animates the ``\textit{leaves}" to flutter.}
    \label{fig:q}
\end{figure}

\noindent \textbf{Quantitative Comparisons.} As shown in Table~\ref{tab:t1}, our method significantly enhances the quality of videos generated by I2V methods AnimateAnything, ConsistI2V, and Cinemo, as well as improves the correspondence between the generated videos and their associated text prompts. 

\noindent \textbf{Qualitative Comparisons.} As shown in Fig.~\ref{fig:q}, when the input is a real painting, the videos generated by AnimateAnything, ConsistI2V, and Cinemo exhibit limitations in preserving image information and imparting motion effects that align with the expectations set by the text prompt. With the addition of our inference method, there is an improvement in the quality of the videos generated by the models. Specifically, AnimateAnything fails to interpret the motion effect intended for the \textit{``boats"} in the text prompt, resulting in a static video. After we introduce synthetic proxy images containing more motion semantics, the generated video features a moving boat as desired. ConsistI2V, while capable of understanding the semantics of the text prompt and animating the \textit{``smoke"} to some extent, produces motion that is not very pronounced. With our method, the generated video shows smoke with more noticeable motion effects and a clearer direction of movement. Cinemo not only fails to understand the semantic information corresponding to the text prompt but also does not preserve the color style of the original image in the generated video. With the addition of our inference method, the generated video retains the original painting's color style while successfully animating the \textit{``leaves”}. For further high-quality video samples, please refer to our project website.

\subsection{Ablation Studies}

\begin{figure}[!t]
    \centering
    \includegraphics[width=0.98\linewidth]{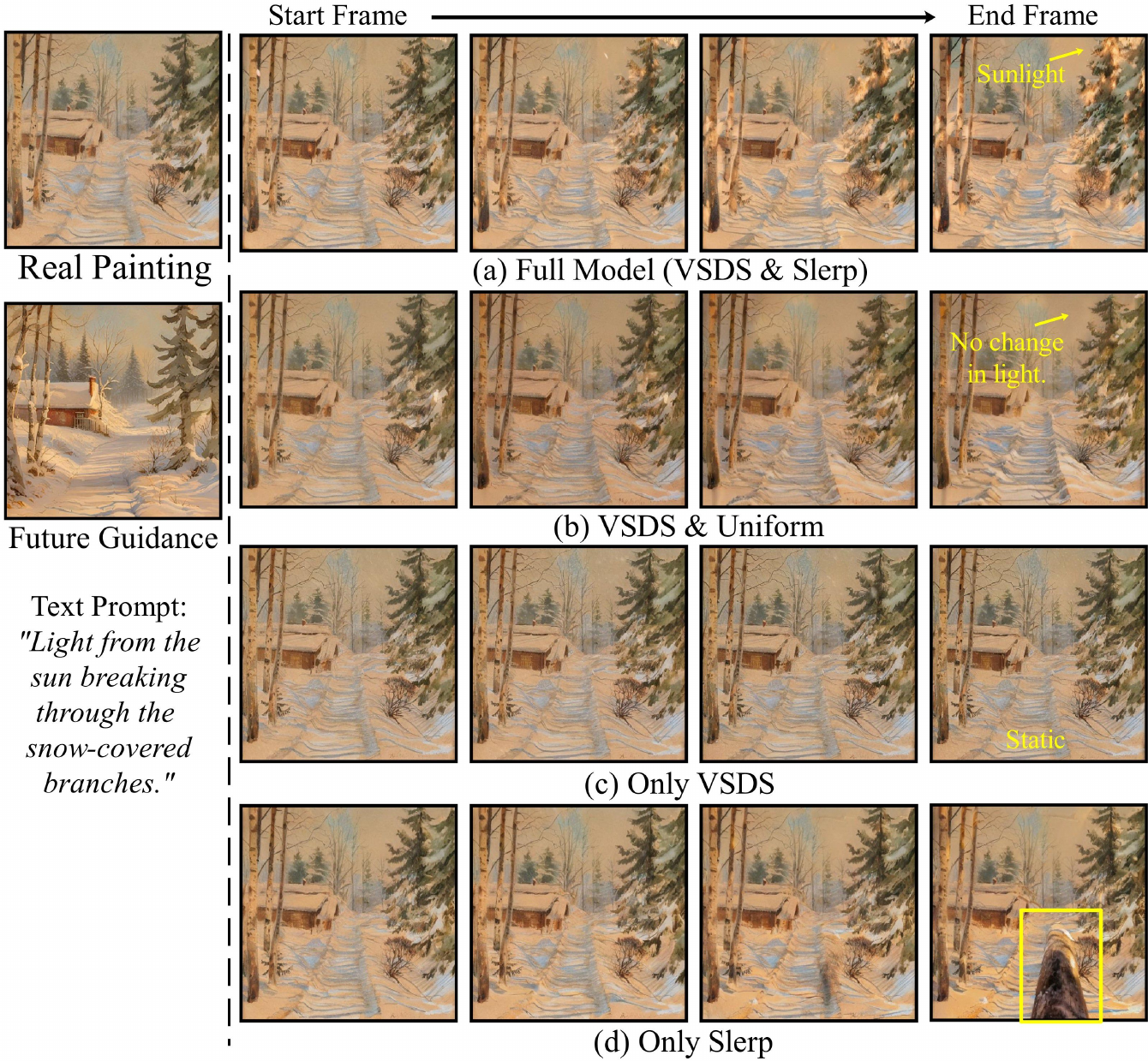}
    \vspace{-0.1in}
    \caption{Ablation study of the two adjustable components. Our full model (a) accurately interprets the semantic content of the text prompt, effectively generating a visual sensation of sunlight penetrating the scene and ensuring that the light source originates from the side with the tree branches. The variant employing uniform linear interpolation (b) fails to adequately capture variations in lighting conditions. The variant that solely performs video score distillation sampling (c) struggles to accurately interpret the semantics of the text prompt, resulting in static video outputs. The variant utilizing only spherical linear interpolation (d) shows erroneous modifications during the latter stages of generation.}
    \label{fig:vas}
\end{figure}

\noindent \textbf{Component Ablation.} Our approach includes two primary components: video score distillation sampling and spherical linear interpolation. To validate the effectiveness of these two components, we design three variant models: one variant replaces spherical linear interpolation with uniform linear interpolation, another variant omits the introduction of synthetic proxy images and only performs video score distillation sampling, and the last variant does not perform video score distillation sampling but instead directly applies spherical linear interpolation between two static video vectors. The quantitative results are shown in Table~\ref{tab:t1}. When uniform linear interpolation is applied instead of spherical linear interpolation, the quality of the generated videos decreases, and there is also a slight reduction in semantic similarity. This indicates that spherical linear interpolation is more suitable for our method. When we do not introduce synthetic proxy images and only add video score distillation sampling to the original inference pipeline of the I2V model, the improvement in semantic similarity of the generated videos is minimal compared to the initial I2V model. This suggests that performing only video score distillation sampling does not significantly enhance the model's ability to understand text prompts. Direct interpolation between two static video vectors results in markedly inferior video quality relative to our complete model. This outcome can be attributed to the fact that the original I2V model is trained using dynamic video vectors, while during inference, the inputs are static video vectors. This mismatch introduces inconsistencies between the training and inference phases. Therefore, it is imperative to initially apply video score distillation sampling to endow the video latent vectors with dynamic characteristics, ensuring a more accurate representation and improved generation quality.

Furthermore, we visualize the videos generated by the three variants in Fig.~\ref{fig:vas}. Our full model accurately understands the semantic information in the text prompt \textit{“Light from the sun breaking through the snow-covered branches”}, generating a sensation of sunlight shining and ensuring the light source originates from the side with the tree branches. The variant employing uniform linear interpolation fails to adequately capture variations in lighting conditions. The variant that solely performs video score distillation sampling struggles with accurately interpreting the semantics of the text prompt, often resulting in static video outputs. Meanwhile, the variant utilizing only spherical linear interpolation exhibits pronounced changes during the latter stages of generation and inadequately preserves the original image information.

\begin{figure}[!t]
    \centering
    \begin{minipage}[b]{0.96\linewidth}
        \centering
        \includegraphics[width=0.85\linewidth]{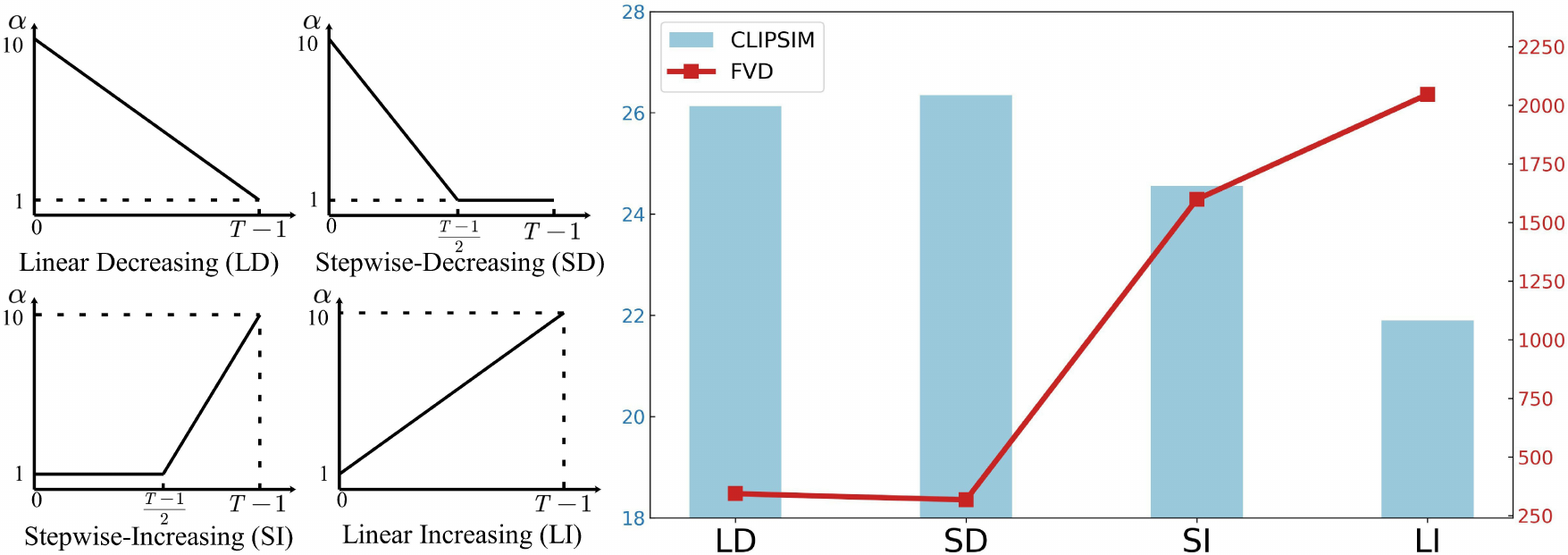}
        \vspace{-0.1in}
        \caption{Ablation study of the weight parameter $\alpha$ in VSDS. Lower FVD values and higher CLIPSIM are better. The model achieves the best performance when the weight curve is Stepwise-Decreasing.}
        \label{fig:pvds_a}
    \end{minipage}
    \begin{minipage}[b]{0.49\linewidth}
        \centering
        \includegraphics[width=1\linewidth]{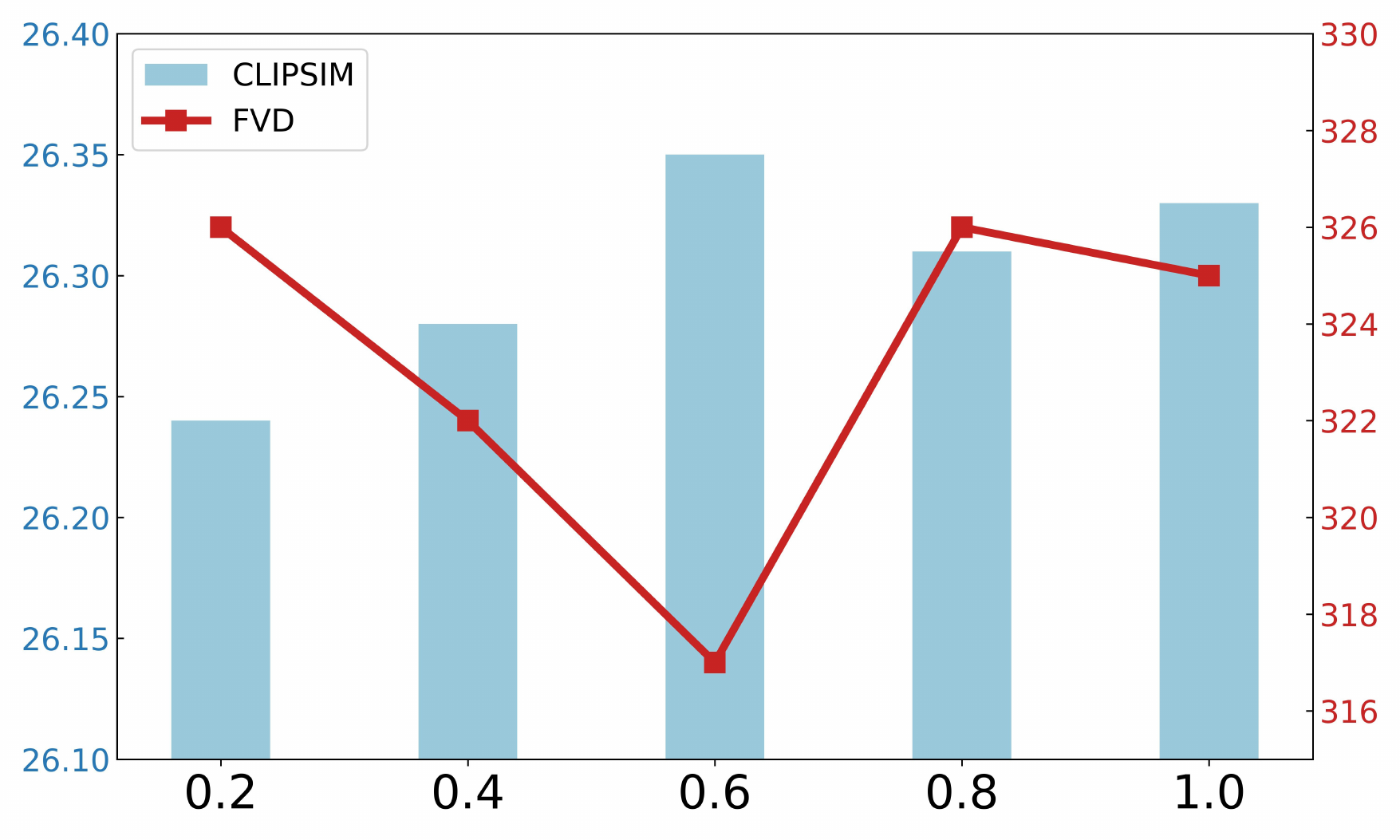}
        \vspace{-0.4in}
        \caption{Ablation study of the time step duration proportion parameter $p$ in VSDS. Lower FVD values and higher CLIPSIM are better. The model achieves the best performance when $p = 0.6$.} 
        \label{fig:pvds_b}
    \end{minipage}
    \begin{minipage}[b]{0.49\linewidth}
        \centering
        \includegraphics[width=1\linewidth]{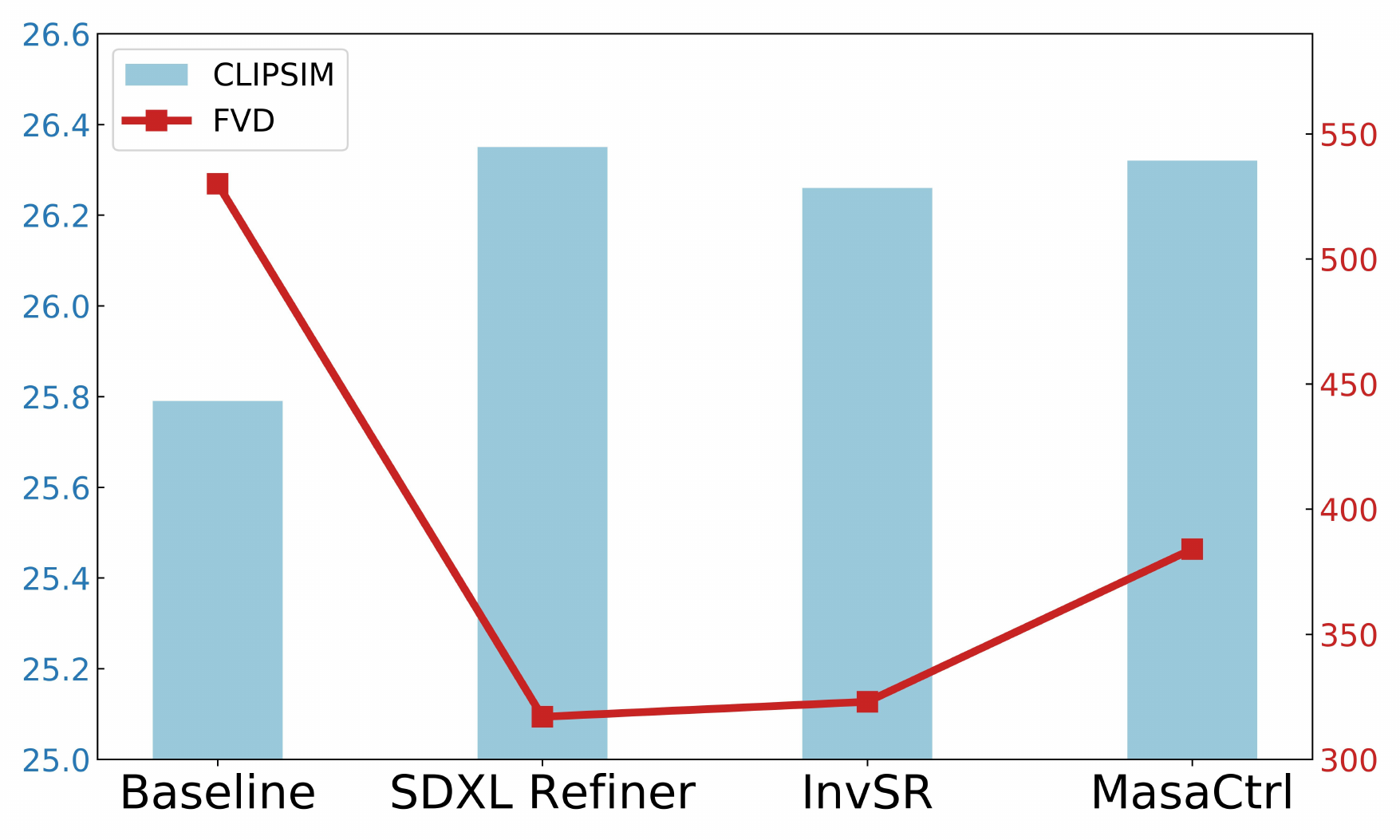}
        \vspace{-0.4in}
        \caption{Quantitative comparison of diverse synthesis strategies. Lower FVD values and higher CLIPSIM are better. Our model with varying proxy images outperforms the baseline model.} 
        \label{fig:i2i_t}
    \end{minipage}
\end{figure}

\noindent \textbf{The parameters of video score distillation sampling.} The video score distillation sampling phase includes two tunable parameters introduced in Eq.~\ref{eq:grad}: the weight $\alpha$ for updating the noise gradient at each time step, and the proportion $p$ that determines the length of the time steps used for updates.

We fix the proportion and vary the updating weights $\alpha$ at different time-steps. The ablation results on $\alpha$ are shown in Fig.~\ref{fig:pvds_a}. We design four sets of weight value curves: Linear Decreasing (LD), Stepwise-Decreasing (SD), Stepwise-Increasing (SI), and Linear Increasing (LI). It can be observed that Stepwise-Decreasing, which employs larger updating weights in the early stages of refinement and maintains an updating weight of 1 in the later stages, achieves the highest semantic similarity and the best video quality. Therefore, we empirically select the settings of Stepwise-Decreasing to configure the updating weights $\alpha$.

We fix the updating weights $\alpha$ and change the proportion $p$ of updating period. As shown in Fig.~\ref{fig:pvds_b}, when the updating proportion $p=0.6$, our method achieves the highest semantic similarity and the lowest FVD. Considering the efficiency of the model and the quality of the generated videos, we set $p=0.6$ for all experiments.

\section{Discussions}
\subsection{Extension to More Natural Images}
\begin{figure}[!t]
    \centering
    \includegraphics[width=1\linewidth]{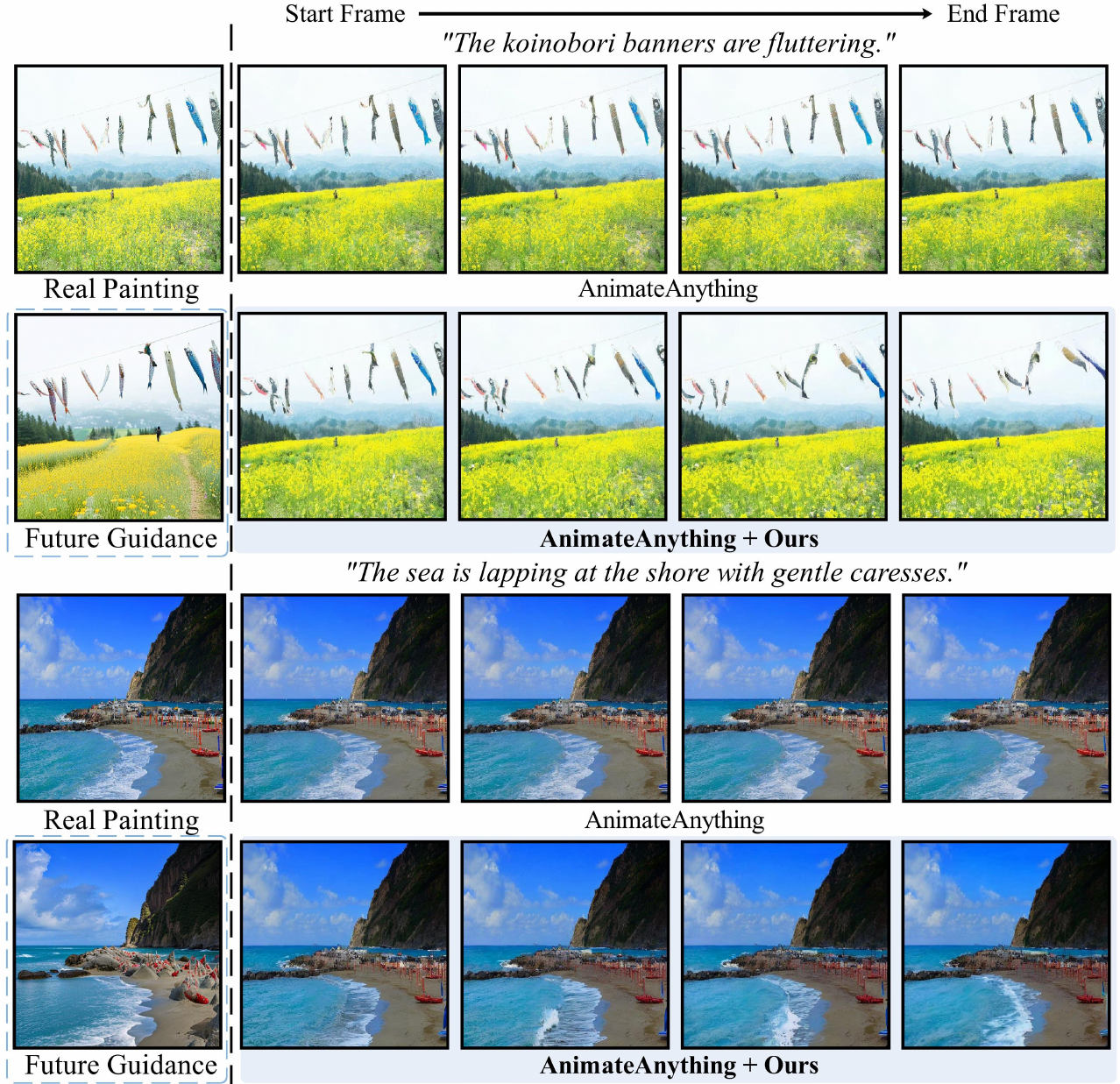}
    \caption{Extension to more natural images. As our baseline model, AnimateAnything fails to adequately understand the meaning of slightly more complex prompts, resulting in videos with weak dynamic effects. However, with the addition of our inference method, the model can more accurately understand the prompt's meaning and generate videos with enhanced dynamic effects.}
    \label{fig:na}
\end{figure}

While our method is principally designed for animating real paintings, it exhibits strong performance when applied to natural images. By capitalizing on the sophisticated semantic understanding capabilities of image models, the introduction of synthetic proxy images significantly improves the I2V model's ability to interpret the text prompt. This enhancement leads to the generation of dynamic effects that more precisely match the intended semantics of the prompt. The results obtained from applying our method to natural images are presented in Fig.~\ref{fig:na}. The baseline model, AnimateAnything, exhibits shortcomings in interpreting the nuances of moderately complex prompts, yielding videos characterized by diminished dynamic qualities. Upon integrating our inference method, the model achieves a notably improved understanding of prompt semantics, which translates to the generation of videos with substantially enhanced dynamic effects.

\subsection{Impact of Diverse Synthesis Strategies}
\begin{figure}[!t]
    \centering
    \includegraphics[width=1\linewidth]{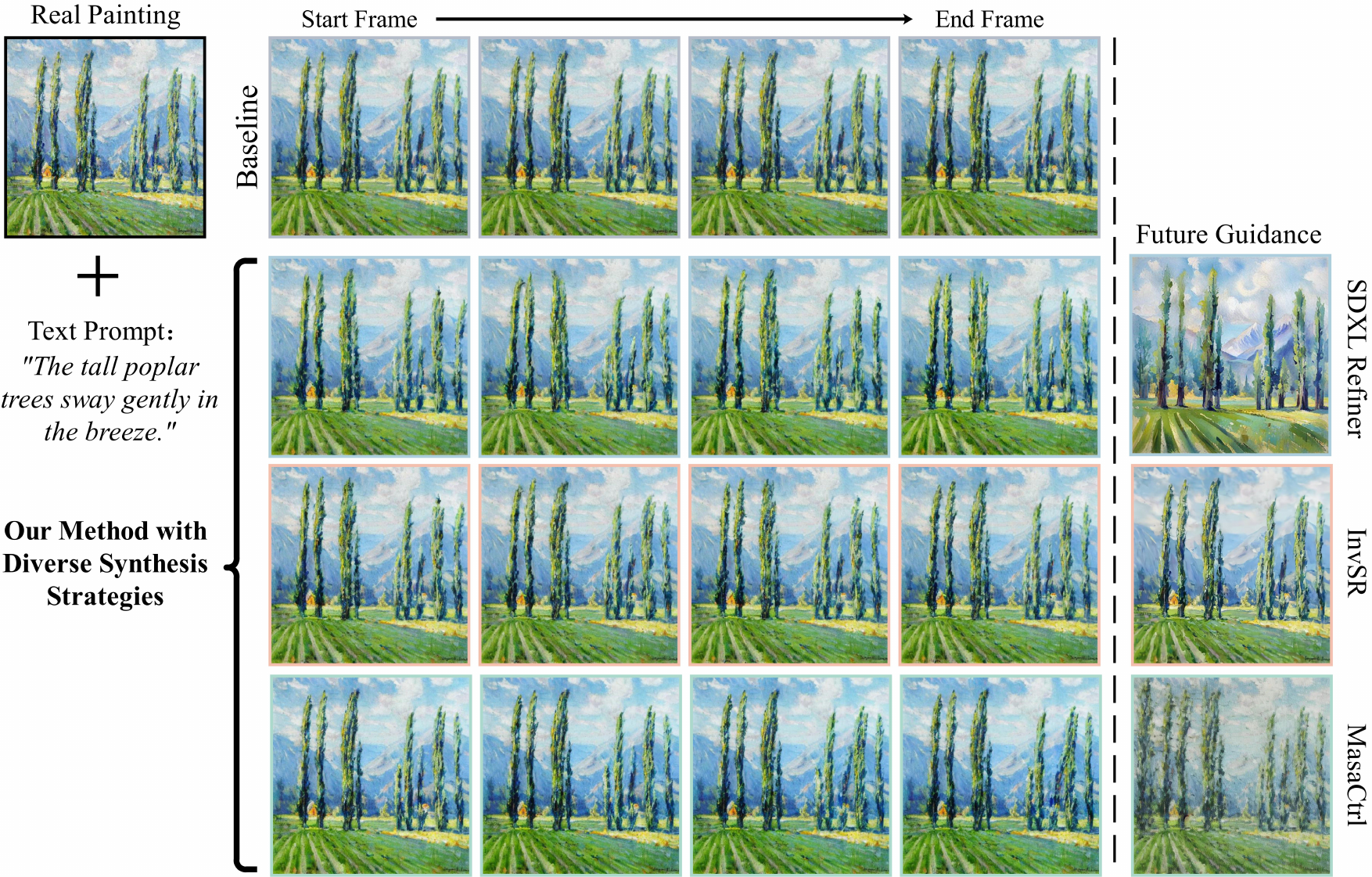}
    \caption{Qualitative comparison of our method with diverse synthesis strategies. The baseline model fails to understand the text instructions and generates a static video. SDXL Refiner (text-driven reconstruction) and InvSR (image-driven reconstruction) produce high-quality images that preserve the original painting's layout and style, enabling coherent motion in the generated videos. MasaCtrl serves as image editing models, introducing subtle but ambiguous edits, resulting in partial motion (\eg, right half of the tree). Even with varying proxy image quality, the videos generated by our method are of higher quality compared to those from the baseline model.}
    \label{fig:i2i}
\end{figure}

We investigate the influence of diverse synthesis strategies on the performance of our method. 
We utilize a fixed real painting and employ three distinct image-to-image models for the generation of synthetic proxy images: SDXL Refiner~\cite{sdxl}, InvSR~\cite{invsr}, and MasaCtrl~\cite{masactrl}. Among these, SDXL Refiner and InvSR serve as image reconstruction models, with SDXL Refiner being text-driven and InvSR being image-driven. MasaCtrl functions as image editing models.
With AnimateAnything as the baseline model, we show the quantitative comparisons in Fig.~\ref{fig:i2i_t}. Across different synthesis strategies, our model consistently outperforms the baseline model.
The qualitative comparisons are presented in Fig.~\ref{fig:i2i}. The images synthesized by SDXL Refiner and InvSR both retain the style of the original painting, with increased resolution and improved image quality. Videos generated using these two synthetic proxy images as inputs can animate according to the expected motion patterns described in the text prompts while preserving the original image's information effectively.
The images generated by MasaCtrl only introduce slight and somewhat ambiguous modifications to the original image. Nevertheless, the corresponding generated videos still manage to animate the trees in the right half. 
Despite variations in proxy image quality, the videos produced by our approach exhibit superior quality relative to those generated by the baseline model.

\subsection{Limitations}
\begin{figure}[!t]
    \centering
    \includegraphics[width=1\linewidth]{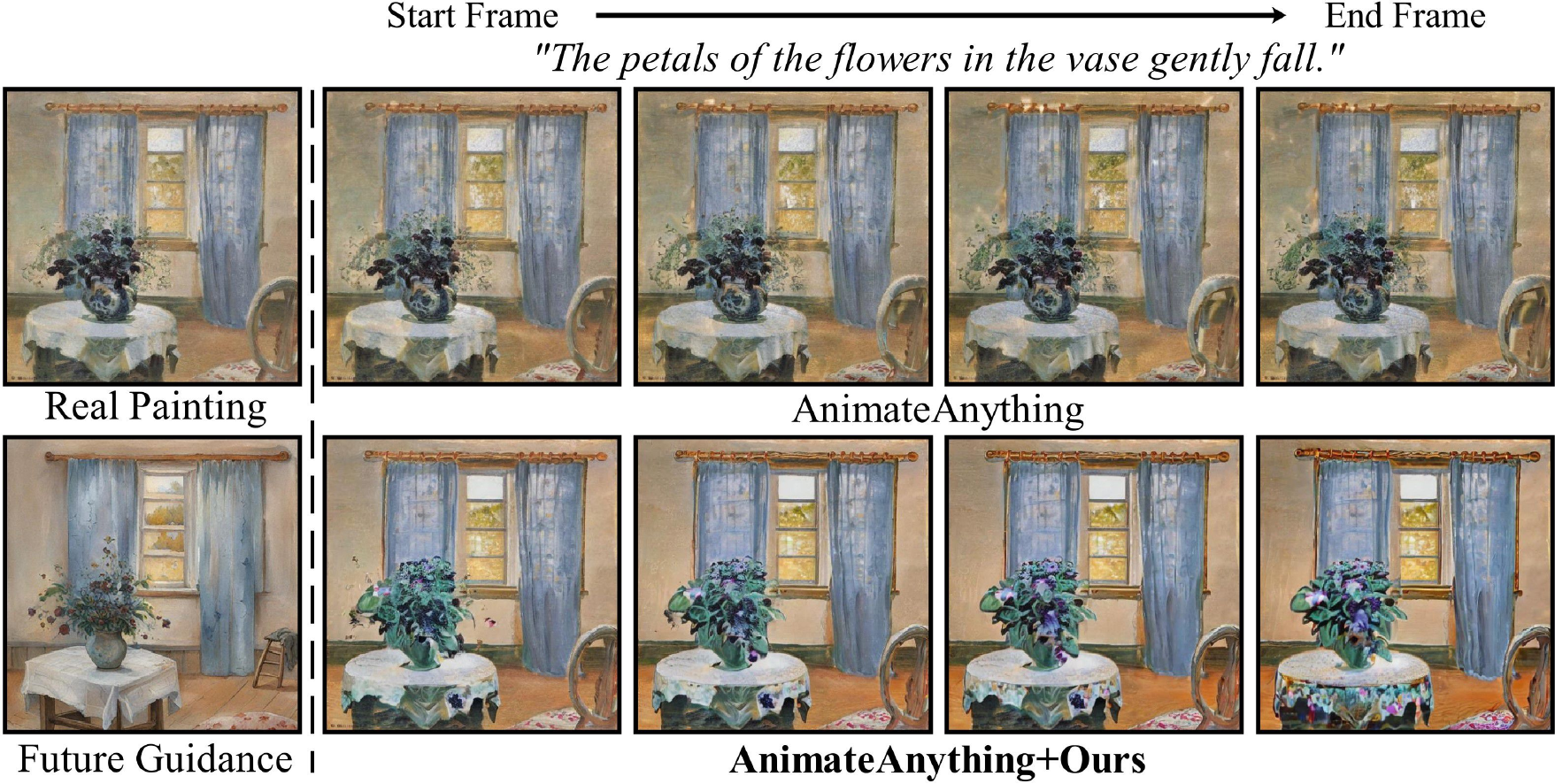}
    \vspace{-0.3in}
    \caption{Failure cases. The baseline I2V model fails to infer complex scene dynamics (\eg, \textit{``falling petals"}), producing static outputs. With synthetic proxy image, the model captures the semantic intent (\textit{``petals falling onto the table"}) but struggles to generate a continuous motion sequence, only partially depicting the process in the final frames due to inherent limitations in temporal coherence.}
    \label{fig:failure}
\end{figure}

Limited by the generative capacities of the baseline I2V model, our method shows deficiencies in generating videos that require intricate scene inference. As shown in the Fig.~\ref{fig:failure}, our baseline model fails to accurately generate a video depicting the imagery of \textit{``falling petals"} thus it tends to produce a static video. After introducing synthetic proxy images, although the model understands the semantic information of the \textit{``falling petals"} as described in the text, its generative capability is limited. Consequently, it only manages to show the imagery of \textit{``petals falling onto the table"} in the final few frames, rather than generating a continuous and complete process of petals falling.

\section{Conclusion}
In this paper, we identify the challenges of animating real paintings via open-domain video diffusion models. We enhance the text responsiveness and video generation quality of I2V models by integrating synthetic proxy images generated by powerful image diffusion models. Specifically, our approach leverages video score distillation sampling and spherical linear interpolation to refine the model's ability to adhere closely to textual descriptions. A key advantage of our method is its plug-and-play nature, eliminating the need for additional learnable parameters or training overhead, thus making it readily applicable to other I2V frameworks without modification. Our solution shows significant improvements in generating high-quality videos from real paintings. Moreover, our approach also exhibits enhancements in generating videos from natural images. Evaluations across three baseline models confirm the competitive performance of our proposed method, highlighting its effectiveness and versatility.

\bibliographystyle{elsarticle-num}

\end{document}